\documentclass[letterpaper, 10 pt, conference]{ieeeconf}  

\IEEEoverridecommandlockouts                              

\usepackage{cite}

\usepackage{amsmath,amssymb,amsfonts}
\usepackage{algorithmic}
\usepackage{graphicx}
\usepackage{textcomp}
\usepackage{xcolor}
\usepackage{multirow}
\usepackage{booktabs}
\usepackage{comment}
\usepackage{dblfloatfix}
\let\labelindent\relax
\usepackage{enumitem}

\usepackage[caption=false]{subfig}
\def\BibTeX{{\rm B\kern-.05em{\sc i\kern-.025em b}\kern-.08em
    T\kern-.1667em\lower.7ex\hbox{E}\kern-.125emX}}

\title{Effects of Explanation Strategies to Resolve Failures  \\in Human-Robot Collaboration
}

\author{Parag Khanna$^{*1}$, Elmira Yadollahi$^{*1}$, Mårten Björkman$^{1}$, Iolanda Leite$^{1}$, and Christian Smith$^{1}$
 \thanks{$^{1}$ Division of Robotics, Perception and Learning (RPL), EECS, KTH Royal Institute of Technology, Sweden
        {\tt\small \{paragk, elmiry, celle, iolanda, ccs\}@kth.se}
        }%
\thanks{* Authors contributed equally}
}

\begin{document}

\maketitle
\thispagestyle{empty}
\pagestyle{empty}

\begin{abstract}
Despite significant improvements in robot capabilities, they are likely to fail in human-robot collaborative tasks due to high unpredictability in human environments and varying human expectations. 
In this work, we explore the role of explanation of failures by a robot in a human-robot collaborative task. 
We present a user study incorporating common failures in collaborative tasks with human assistance to resolve the failure. 
In the study, a robot and a human work together to fill a shelf with objects. 
Upon encountering a failure, the robot explains the failure and the resolution to overcome the failure, either through handovers or humans completing the task. 
The study is conducted using different levels of robotic explanation based on the failure action, failure cause, and action history, and different strategies in providing the explanation over the course of repeated interaction. 
Our results show that the success in resolving the failures is not only a function of the level of explanation but also the type of failures.  
Furthermore, while novice users rate the robot higher overall in terms of their satisfaction with the explanation, their satisfaction is not only a function of the robot's explanation level at a certain round but also the prior information they received from the robot. 
\end{abstract}

\section{Introduction}


Robots and artificial agents' capabilities rapidly grow as they are deployed in real-world environments like factories, hospitals, and schools. 
Nevertheless, failures inevitably occur during task execution and collaboration \cite{van2022correct}, and with the increasing use of robots in in-the-wild environments, where robots are more prone to collaborate with novice and non-expert users, the study of failures, and mitigating their impact becomes imminent.
While in many failure scenarios, robots can recover by themselves, there are cases where human assistance is required to resolve failures for task continuity \cite{bauer2008human}.
For a non-expert user, understanding \textit{why} a robot failure has occurred and if and \textit{how} they could contribute to the recovery is essential for smooth human-robot collaboration. 

The emergence of studies on robot failures attests to the evolution of research from exploring people's perception and resolution of failures to the robot's role in identifying and mitigating them \cite{honig2018understanding}.
While the topic has expanded to include the use of holistic approaches such as explanation, apology, denial, and promise, that identify, resolve, and mitigate failures for untrained users \cite{van2022correct, das2021explainable, karli2023if}, few have studied the effect of these approaches in repeated interactions to the best of our knowledge \cite{esterwood2023three}. 
Providing an explanation is a practical approach to mitigating failures in collaborative scenarios, particularly when failures require human intervention or assistance. 
Advances in the field of Explainable AI (XAI) \cite{gunning2019darpa}, and its extension to goal-driven explanations \cite{sakai2022explainable} for robots and agents contribute to research on explanation generation for failures. 
Currently, the literature on the topic of resolving failures via explanations focuses on determining \textit{what} type of information should be presented \cite{das2021explainable} and \textit{how} the explanations should be automatically generated \cite{eder2022fast, diehl2022did}.
In our research, we address the missing link between explanation generation and participant satisfaction in repeated interactions with recurring failures, for example, do we need to be consistent with the explanations as the failures reoccur, or should we provide more details early on and reduce as the interaction continues? 

\begin{figure}[t]
      \centering
        \includegraphics[width=.85\linewidth,trim={6.0cm 2.0cm 3.5cm 2.8cm},clip]{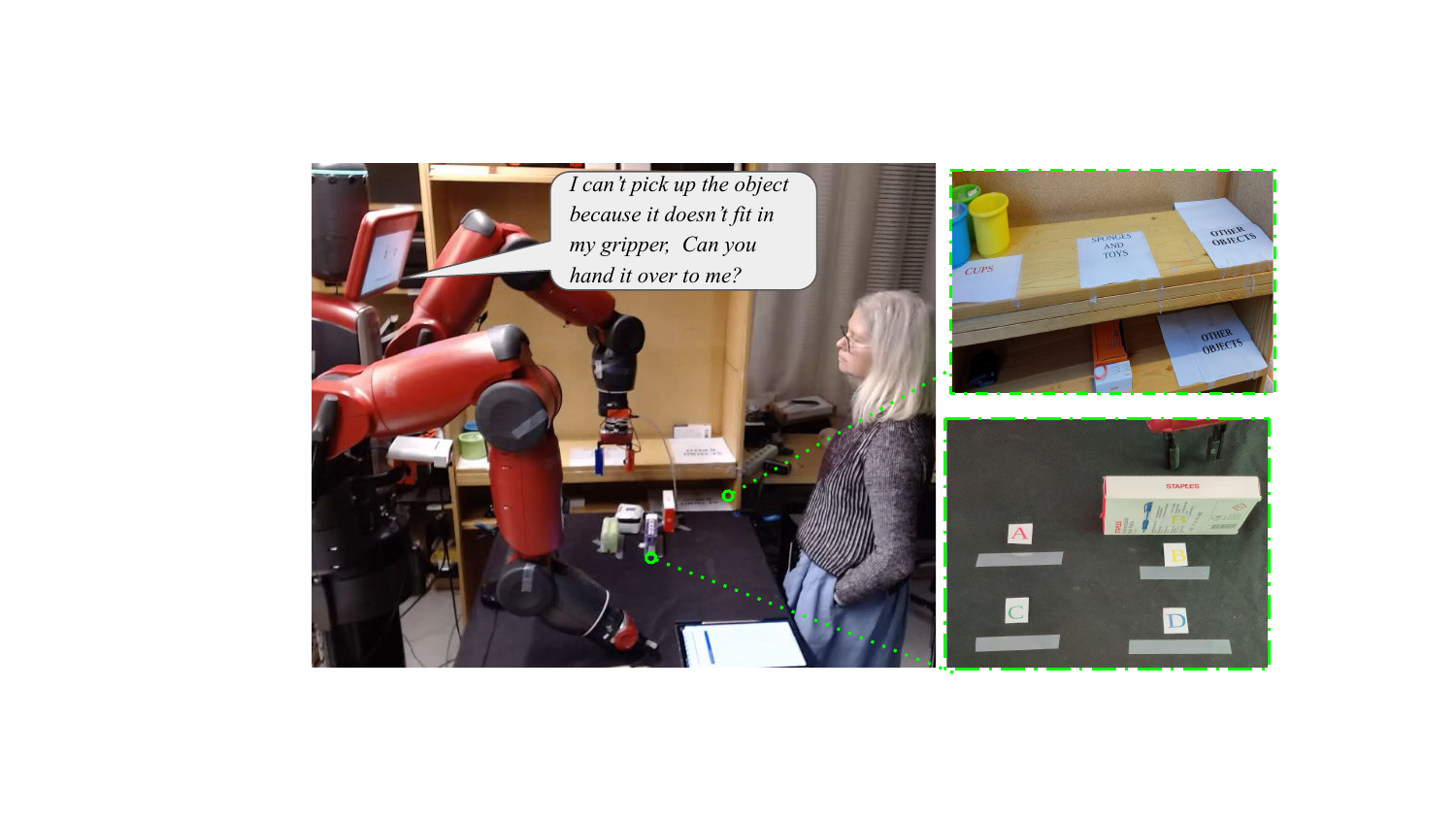}  
    \setlength\abovecaptionskip{-0.8\baselineskip}
    \caption{The pick and place task: Human places the objects on the table, then the robot's goal is to place them on the shelf while providing an explanation in case a failure occurs. The zoomed views show (top right) two levels of the shelf and (bottom right) the markers for object placement on the table.}
      \label{fig:HRC_task_view}
\end{figure}

\begin{figure*}[hb]
\captionsetup[subfigure]{justification=centering}
\centering
 \subfloat[Sponge (10 g) and (left) and Toy (25 g) (right)][Sponge 10g (left), \\ Toy 25g (right)]{
\includegraphics[width=.18\linewidth]{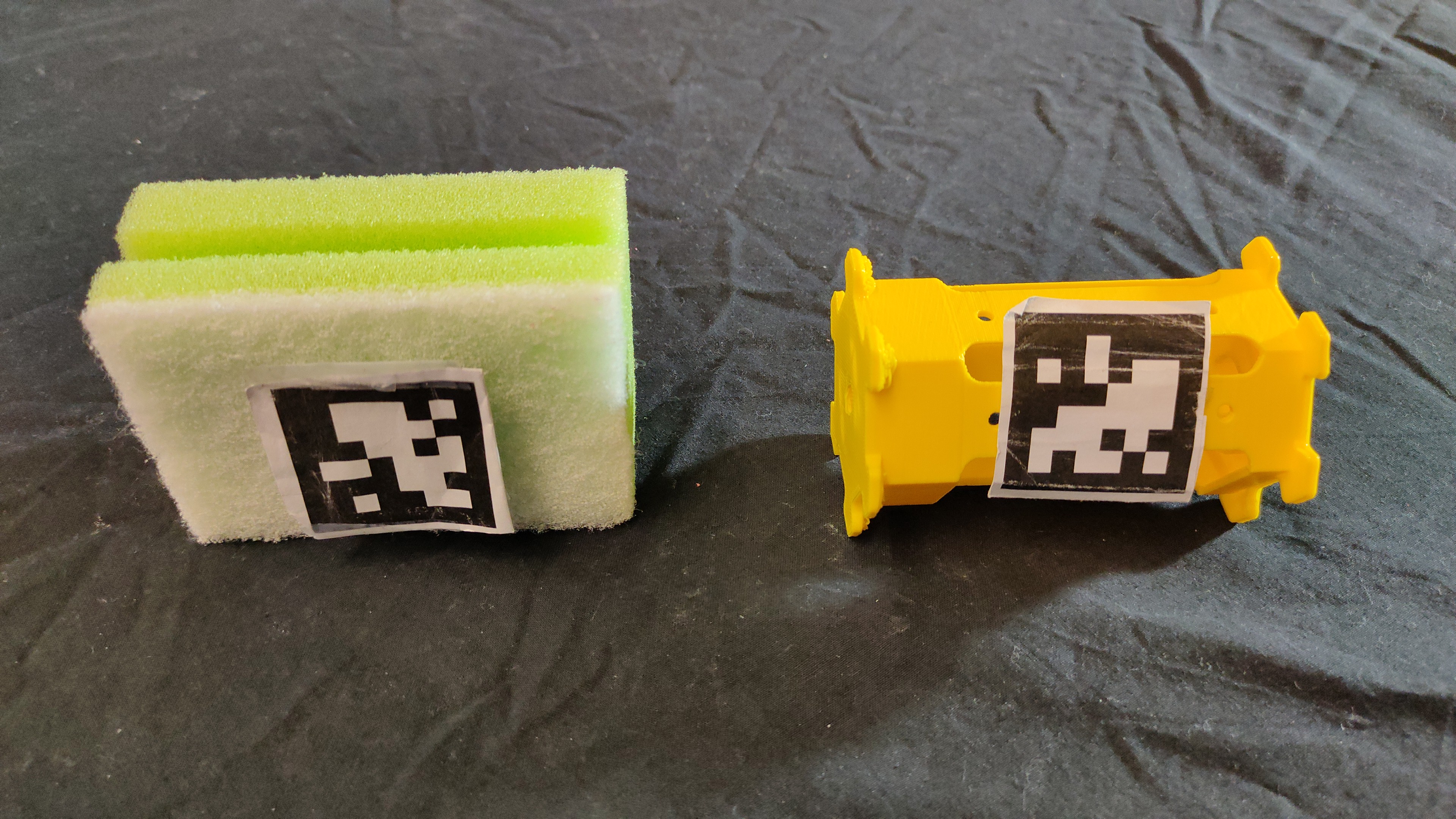}
}
\subfloat[Pen box (left) and Pen box-heavy (right)][Pen box 500g (left), \\ Pen box-heavy 750g (right)]{
\includegraphics[width=.18\linewidth]{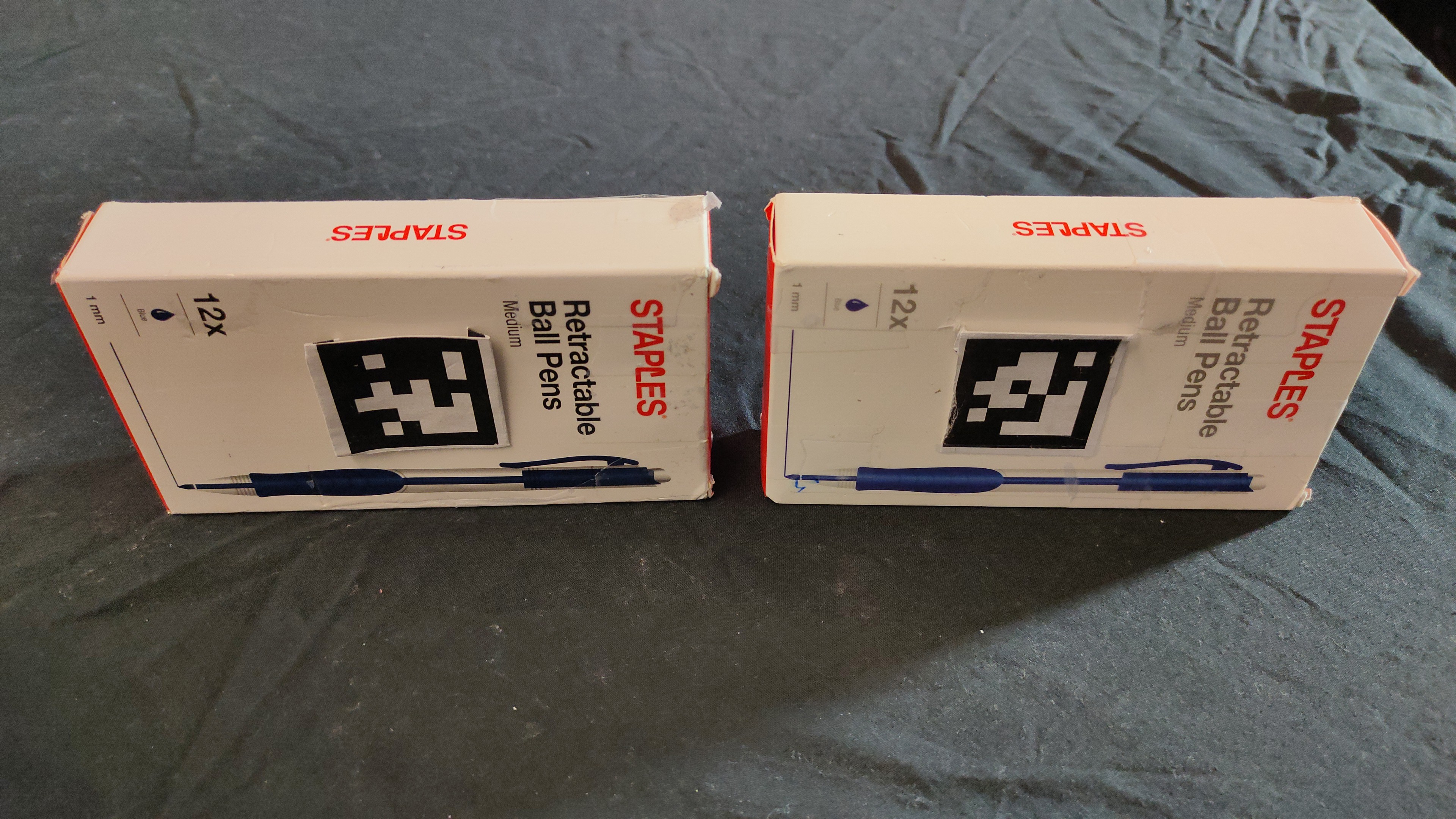}
}
\subfloat[Heavy box 1 725g (left), Heavy box 2 1000g(right)][Heavy box 1 725g (left), \\ Heavy box 2 1000g (right)]{
\includegraphics[width=.18\linewidth]{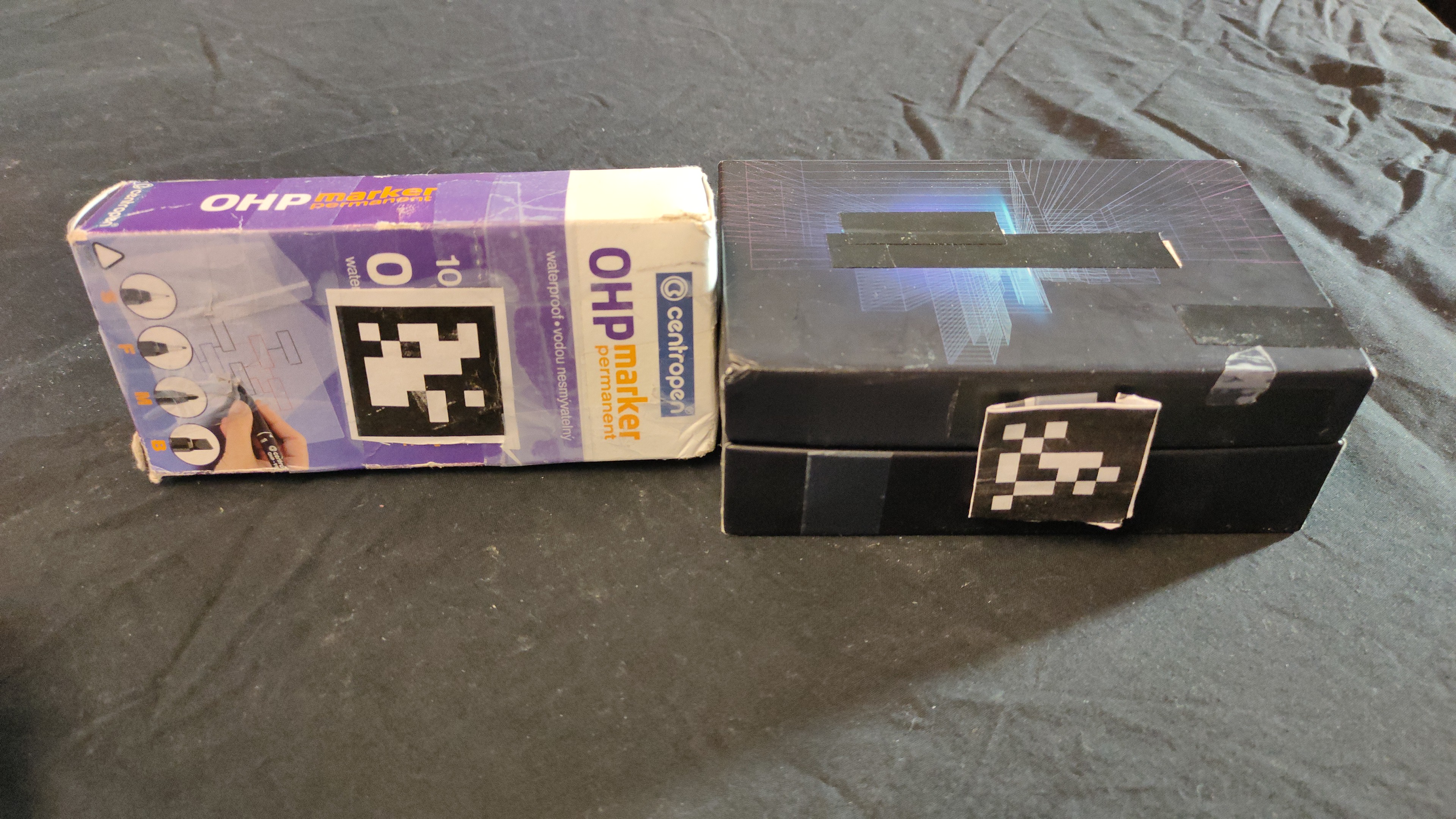}
}
\subfloat[Cloth box 710g, Flat box 130g][Cloth box 710g, \\ Flat box 130g]{
\includegraphics[width=.18\linewidth]{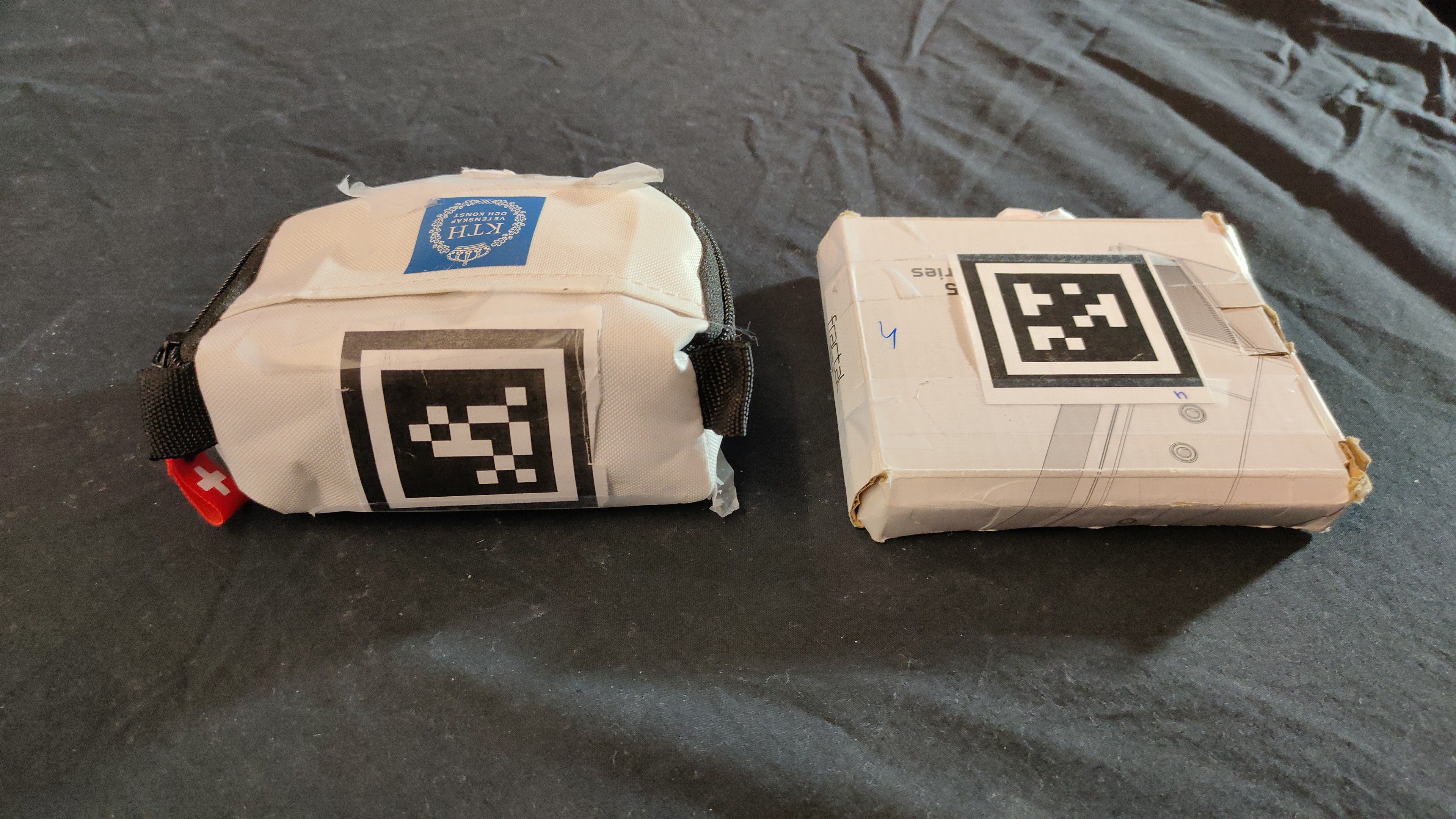}
}
\subfloat[Random Box 2 (left), Random Box 1 (right)][Random Box 2 580g (left) \\ Random Box 1-3 35g (right)]{
\includegraphics[width=.18\linewidth]{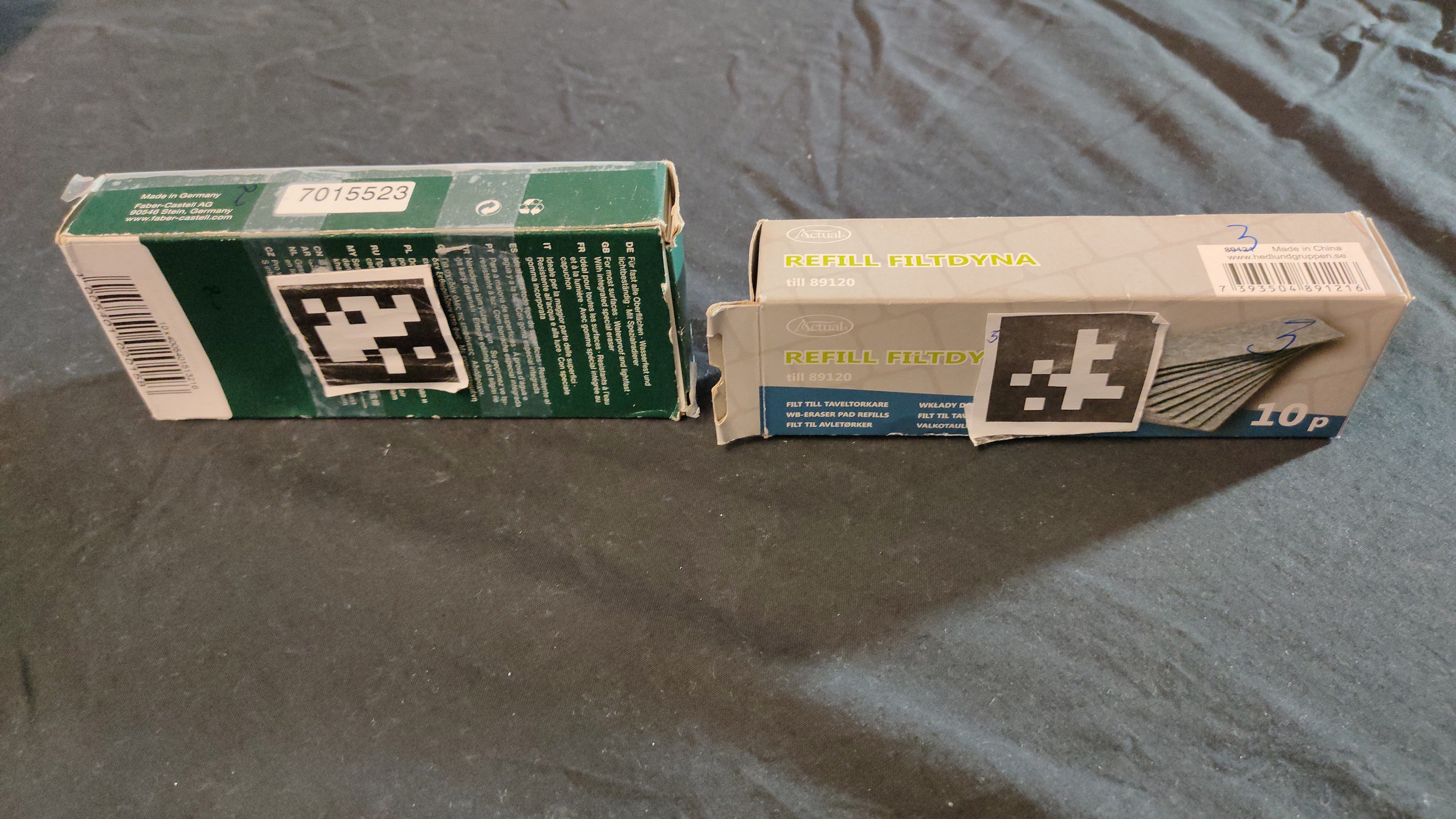}
}
\caption{Objects as they were required to be placed in front of the robot}
\label{fig:Objects}
\end{figure*}


As a result, we developed a study to understand how different strategies of providing explanations in repeated interaction influence non-expert users' performance and satisfaction. 
We developed a collaborative pick-and-place task, where the robot and human had to place objects from four baskets on a shelf. 
We counted each basket as one round of interaction and aimed to have four rounds of interaction. 
We designed two types of strategies for providing the explanations: 1) maintaining the details of the explanation during the rounds, i.e. \emph{fixed strategy} and 2) reducing the details of the explanation, i.e. \emph{decaying strategy}. 
To develop the strategies, we first defined the explanation levels inspired by the previous work by \cite{das2021explainable} and labeled them as low, mid, and high.
Subsequently, we conducted a between-subject user study where participants experienced either of these strategies in four rounds of interaction. 
We aimed at evaluating how participants' performance in resolving failures and satisfaction with the explanations were impacted by the explanation levels and strategies which lead us to the following research questions:

\begin{itemize}[leftmargin=*]
    \item \textit{RQ1: How does explanation level impact participants' performance in the task and satisfaction with the explanation?} 
\end{itemize}

\begin{itemize}[leftmargin=*]
    \item \textit{RQ2: Which explanation strategy (fixed vs. decaying) leads to better performance and satisfaction in participants?} 
\end{itemize}

\section{Related Work}

Several studies in the fields of human-robot interaction (HRI) and collaboration (HRC) have addressed the importance of understanding the effect of failures on trust \cite{Correia_fault_justification} and perception \cite{van2017lack}, and mitigating its impact through failure recovery \cite{reig2021flailing}, explainability \cite{das2021explainable}, and promise \cite{karli2023if}. 
In \cite{Correia_fault_justification}, a user study was designed to investigate the effects of a collaborative robot's failure on human trust and the impact of justification strategies. 
Altogether, the results indicated that a faulty robot is regarded as far less trustworthy. 
It is also shown that the impact of failures was reduced with justifications when the consequence of failure was less significant.
With the change of trend in using holistic approaches to identify and resolve failures, one of the approaches used more commonly in recent years is generating and providing explanations, studied on both the computational front, e.g., XAI \cite{alvanpour2020robot, das2021explainable}, and the social front, e.g., behavioral \cite{miller2021contrastive}. 
Several of the research in explainability has been inspired by the sociocognitive definitions of explanations in various fields and their social implications \cite{miller2021contrastive}.
A recent review by Wallkotter et al. \cite{wallkotter2021explainable} identified three research directions on the topic that contribute to understanding the explainability mechanisms and how they can be integrated into the interaction context with occasional overlaps with the field of XAI. 
On the topic of studying explainability in robot failures, a study by Das et. al. investigated the types of explanation that helped non-experts to identify robot failures and assist the recovery by extending the XAIP algorithms via introducing failure explanation \cite{das2021explainable}.
The goal was to produce explanations for unexpected failures in a pick-and-place task for a robot in a household environment. 
Failure and solution identification has been observed to be most effective when explanations include the context of the failure action and the history of previous actions. 
Another study in \cite{alvanpour2020robot} used machine-learning models to predict robot grasp failures and study the tradeoff between accuracy using black-box models and interpretability using explainable models. 
They showed an explanation of predicted faults could contribute to the efficiency of designing the robot and avoiding future failures.
Diel et al. proposed a causal-based method to develop explanations for robot failures in collaborative scenarios \cite{diehl2022did}.
Their approach incorporated learning from a causal Bayesian network that enabled the robot to generate 
the explanation by contrasting a failure state against the closest successful state and by using a breadth-first search.
Beyond the studies focusing on generating explanation, the effects of different types and amounts of explanation by an XAI system on human understanding of the system were discussed in \cite{XAI-Linder_effects_Exp_levels}, where an increase in the information contained in the explanation resulted in the users' better understanding and prediction of the system behavior, as well as increased user performance. However, this came at a cost of increased time and attention needed by users to comprehend the explanation.

\section{Design}

\subsection{Collaborative Task Design}
\def\mystrut{\rule{0pt}{0.80\normalbaselineskip}}
\begin{table*}[hb]
\setlength\abovecaptionskip{-0.35\baselineskip}
\caption{Failure-wise Explanation levels}
  \centering
  \setlength{\tabcolsep}{2.5pt}

  \begin{tabular}{@{}p{1.3cm}p{3.1cm}p{3.1cm}p{3.1cm}p{3.1cm}p{3.0cm}@{}}
    \toprule
    \textbf{Level} & \multicolumn{1}{c}{\textbf{$f_0$}} &  \multicolumn{1}{c}{\textbf{ $f_1$ }} & \multicolumn{1}{c}{\textbf{ $f_2$ }} & \multicolumn{1}{c}{\textbf{ $f_3$ }}  & \multicolumn{1}{c}{\textbf{ $f_4$ }} \\
    \midrule
    \textbf{Zero} (Non-verbal) & Shakes its head to show unable to find the object & Tries to get the object in its gripper, shakes its head at failure, and moves the arm to the handover position & Moves the arm down with the object in the gripper, conveying it fails to carry it, shakes head and moves to the handover position & Stops arm near the lower level of the shelf, shakes head and move the arm to the handover position & Shakes head \\\mystrut
    Resolution & Nothing & Nothing & Nothing & Nothing & Nothing \\
     \midrule
    \textbf{Low} & ``I can't detect the object" & ``I can't pick up the object" & ``I can't carry the object" & ``I can't place the object" & ``I failed to handle the object" \\\mystrut
    Resolution & ``Move it" & ``Hand it to me" & ``Carry it for me" & ``Place it for me" & ``Place it for me"\\
    \midrule
    \textbf{Medium} & ``I can't detect the object because the tag is not visible to me" & ``I can't pick up the object because it doesn't fit in my gripper" & ``I can't carry the object because it is too heavy for my arm" & ``I can't place the object because the destination is out of my arm's reach" & ``I failed to handle the object because an unexpected failure happened" \\\mystrut
    Resolution & ``Can you move it within my field of view?" & ``Can you hand it over to me?" & ``Can you carry the object for me?" & ``Can you place the object for me?" & ``Can you place the object for me?"\\
    \midrule
    \textbf{High} & ``I scanned all the objects and can't detect the object I am looking for, because probably the tag is not visible to me" & ``I can detect the object, but I can't pick it up because it doesn't fit in my gripper" & ``I can pick up the object, but I can't carry it because it is too heavy for my arm" & ``I can carry the object, but I can't place it because the destination is out of my arm's reach" & ``I can detect the object but I failed to handle it because an unexpected failure happened"\\\mystrut
    Resolution & ``Can you move it within my field of view to make sure I see the tag?" & ``Can you hand it over to me by placing it in my gripper?" & ``Can you carry the object for me and place it on the shelf?" & ``Can you place the object on the shelf location that is out of my reach?" & ``Can you finish placing the object for me?"\\
    \bottomrule
  \end{tabular}
\label{tab:exp_levels}
\end{table*}
We designed a pick-and-place task where a Baxter robot  and a human had the goal of collaboratively picking objects from a basket and placing them on the shelf (Fig. \ref{fig:HRC_task_view}).
We created four baskets, (numbered 1 to 4), each including a combination of four household items presented in Fig \ref{fig:Objects}. 
This resulted in a total of 16 objects that needed to be placed on the shelf during the whole duration of the experiment. 
In our design, each round of the experiment started by picking the items from one basket, putting them on the table, and placing them on the shelf, when the task was successfully executed. The placement of an object was deemed unsuccessful if it was not placed on the shelf. 

We marked each object in the basket with an A, B, C, or D tag on one face and a fiducial tag \cite{Apriltag_olson2011tags} on the other to let the robot detect the object.
At the start of each round, the human collaborator placed all objects from the basket in corresponding positions as they are marked (see Fig. \ref{fig:HRC_task_view}).
For handling each object, the robot executed the following steps: \textit{detect} the object, \textit{pick} it up, \textit{carry} it, and finally \textit{place} it on the shelf. 
A possible failure could happen at each step during collaboration with the robot.
As result, we defined the following failures and possible resolutions that could help complete the task despite a failure. 
In the next section, the explanations generated based on these failures and resolutions are provided. 
\begin{enumerate}[leftmargin=*]
    \item \textbf{Detect Failure ($f_0$):} Robot failed to detect the object on the table, e.g. not being able to scan the tag.
    
    \textbf{Resolution Action ($r_0$)} Human moves or rotate the object to ensure the tag is visible to the robot.
    
    \item \textbf{Pick Failure ($f_1$)}: Robot failed to pick up an object, e.g. not fitting in the gripper, based on its placement or size.
    
    \textbf{Resolution Action ($r_1$)}: Human picks up and hands over the object to the robot. 
    
        \item \textbf{Carry Failure ($f_2$)}: Robot failed to carry an object, e.g. weight beyond the limit robot can handle.
    
    \textbf{Resolution Action ($r_2$)}: Robot hands over the object to the human and the human places it on the shelf. 
    
    \item \textbf{Place Failure ($f_3$)}: Robot failed to place an object, e.g. the desired destination is beyond the robot's reach.
    
    \textbf{Resolution Action ($r_3$)}: Robot hands over the object to the human, and they place it at the desired location.
\end{enumerate}

\begin{figure}[t]
\vspace{2mm}
      \centering
        \includegraphics[width=.77\linewidth,trim={3.6cm 0.0cm 1.8cm 0.0cm},clip]{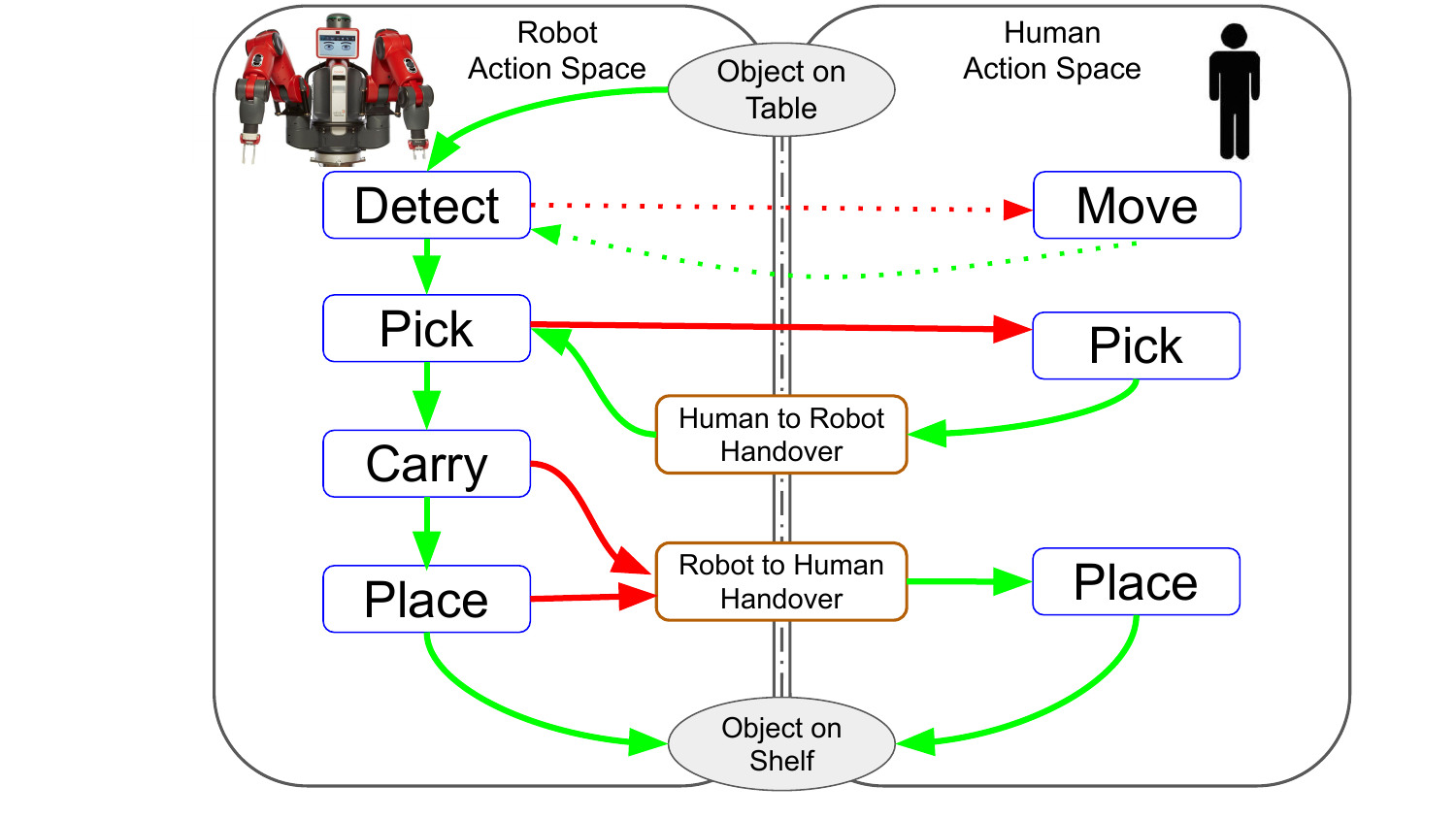}  
    \setlength\abovecaptionskip{-0.5\baselineskip}
    \caption{Description of human-robot collaborative task with the robot and human action spaces. Arrows in green represent transitions due to action success. Arrow in red represents transitions due to action failure. }
      \label{fig:HRC_collab_task}
\vspace{-2mm}
\end{figure}

Fig. \ref{fig:HRC_collab_task} shows the workflow for placement of an object with possible failures denoted in red, which was accompanied by an explanation from the robot and resolved from the human side. In the task design, we included 7 objects for which the robot successfully executes all steps and 9 objects involving some robotic failures, spread out across the four rounds as shown in Table \ref{tab:HRC_round_objects_1}. 
We are not intentionally incorporating the detection failures ($f_0$), but as they might occur due to the way the object is placed on the table, we provide the appropriate resolution. 
For any other unintended failure ($f_4$), the resolution ($r_4$) in the form of asking the human to place the object on the shelf was also integrated.
\begin{table}[t]
\centering
\setlength\abovecaptionskip{-0.4\baselineskip}
\caption{Round-wise Description}

\resizebox{0.9\columnwidth}{!}{
    \begin{tabular}{c c c c c c c}
         \toprule
     Round & Object & Object &  \multicolumn{3}{c}{Robotic Action-Success} & Resolution \\ 
     \cmidrule(lr){4-6} 
        No. & & Type & Pick & Carry & Place & \\
        \midrule
        \multirow{4}{*}{1}& A & Sponge & \textbf{\checkmark} & \textbf{\checkmark} & \textbf{\checkmark} & None\\ 
        & B & Cloth-Bag & \text{\sffamily X}& \text{\sffamily X} & - & $r_1$ \& $r_2$ \\
        & C & Random-Box1 & \textbf{\checkmark} & \textbf{\checkmark} & \textbf{\checkmark} & None\\
        & D & Pen-Box & \textbf{\checkmark} & \textbf{\checkmark} & \text{\sffamily X} & $r_3$ \\ 
        \midrule 
        \multirow{4}{*}{2}& A & Heavy-Box1 & \textbf{\checkmark} & \text{\sffamily X} & - & $r_2$\\ 
        & B & Random-Box2  & \textbf{\checkmark} & \textbf{\checkmark} & \textbf{\checkmark} & None\\
        & C & Flat-Box & \text{\sffamily X} & \textbf{\checkmark} & \textbf{\checkmark} & $r_1$\\
        & D & Toy & \textbf{\checkmark} & \textbf{\checkmark} & \textbf{\checkmark} & None\\ 
        \midrule
        \multirow{4}{*}{3}& A & Pen-Box & \textbf{\checkmark} & \textbf{\checkmark} & \text{\sffamily X} & $r_3$\\ 
        & B & Pen-Box-Heavy & \textbf{\checkmark} & \text{\sffamily X} & - & $r_2$\\
        & C & Toy & \textbf{\checkmark} & \textbf{\checkmark} & \textbf{\checkmark} & None\\
        & D & Random-Box3 & \textbf{\checkmark} & \textbf{\checkmark} & \textbf{\checkmark} & None\\ 
        \midrule
        \multirow{4}{*}{4}& A & Flat-Box & \text{\sffamily X} & \textbf{\checkmark} & \textbf{\checkmark} & $r_1$\\ 
        & B & Sponge & \textbf{\checkmark} & \textbf{\checkmark} & \textbf{\checkmark} & None\\
        & C & Pen-Box & \textbf{\checkmark} & \textbf{\checkmark} & \text{\sffamily X} & $r_3$\\
        & D & Heavy-Box2 & \text{\sffamily X} & \text{\sffamily X} & - & $r_1$ \& $r_2$\\ 
        \bottomrule
    \end{tabular}
}
\label{tab:HRC_round_objects_1}
\vspace{-2mm}
\end{table}
\subsection{Explainability Mechanisms}
We considered three verbal explanation levels: low, medium, and high. 
Additionally, we included a nonverbal baseline to complement the explanations as a result of initial pilot studies where we noticed users needed some baseline behaviors to understand the failures, particularly when given low-level explanations.
As a result, we designed the following explanation levels inspired by \cite{das2021explainable} and Table \ref{tab:exp_levels} presents each explanation for each failure type.

\begin{itemize}[leftmargin=*] 

    \item Low Level: Based on \textit{action-based} explanation in \cite{chakraborti2020emerging}. After the failure, the robot states the failure action and its resolution.
    
    \item Medium Level: Based on \textit{context-based} explanation in \cite{chakraborti2020emerging}. Post failure, the robot states the failed action and the cause of failure, followed by a resolution statement.
    
    \item High Level: Based on \textit{context-based + history-based} explanation in \cite{chakraborti2020emerging}. After failure, the robot states the previous successfully completed action, the current failure action, and its cause. The resolution statement also includes the resolution action.
\end{itemize}

Informed by our pilot studies, we included a nonverbal baseline to help with identifying the failure in lower explanation levels.

\begin{itemize}[leftmargin=*] 
    \item Zero (Nonverbal): This only includes the robot head shaking at each failure with more specific robotic actions based on the failure type.
\end{itemize}


\subsection{Interaction Details}
The Baxter robot was programmed in ROS and only used its left arm. 
More detail on the technical developments and the interaction is available in \cite{khanna_study_wshop_HRI} and the accompanying video with this work.
Each round started with the robot receiving verbal confirmation that all objects are placed on the table, where the robot proceeded to pick up the objects by following the action sequence depicted in Fig. \ref{fig:HRC_collab_task}.
Once a failure occurred, the robot exhibited non-verbal actions described in Table \ref{tab:exp_levels}. 
followed by an explanation based on the current strategy and waiting for the human to resolve the failure before moving to the next step.
If the failure was not resolved in a predefined amount of time, the robot repeated itself up to five times spaced with three-second intervals. 
The system is autonomous, but the experimenter (unbeknown to the participant) made the decision to move to the next step when they failed to complete the task after five repetitions (something that might happen in low explanation cases).
To avoid handover failures, the human-to-robot handover was completed after the robot received a verbal confirmation to close its gripper after the human handed over the object, and the robot-to-human handover used sufficient pull-force, in line with a prior study \cite{parag-humanoids}.

\section{Methodology}

\subsection{Experiment Design}

We investigated two explanation strategies (fixed and decaying) using the three levels of explanations (low, mid, and high). 
For the fixed explanation strategy, we tested each explanation level using the three conditions: C1, C2, and C3 presented in Table \ref{tab:exp_cond}. 
For the decaying explanation strategy, we focused on the rate of decay.
Given four rounds of interactions, we defined two types of decay: \textit{slow} (D1) and \textit{rapid} (D2).
Slow decay was implemented by reducing the level of explanation once per round, which resulted in the following combination: high, medium, low, and none. 
In Rapid decay, the  explanation was reduced from high to low and keep it in a low level as presented in Table \ref{tab:exp_cond}.

\begin{table}[t]
\setlength\abovecaptionskip{-0.4\baselineskip}
\caption{Experimental Conditions}
\centering
\label{tab:exp_cond}
\resizebox{0.965\columnwidth}{!}{

\begin{tabular}{llllll}
\toprule
ID  & Details & Round 1 & Round 2 & Round 3  & Round 4          \\\midrule
\textbf{C1} & Fixed-Low         & Low       & Low       & Low       & Low           \\
\textbf{C2} & Fixed-Medium      & Mid       & Mid       & Mid       & Mid           \\
\textbf{C3} & Fixed-High        & High       & High       & High       & High            \\
\textbf{D1} & Decay-Slow        & High       & Mid       & Low       & None           \\
\textbf{D2} & Decay-Rapid      & High       & Low       & Low       & Low  \\
\hline
\end{tabular}
}
\vspace{-1mm}
\end{table}

\subsection{Hypotheses}
Prior research on the topic of XAI and explainability in robotics has shown mixed results in how humans perceive explanations. 
In the study by Das et. al. \cite{das2021explainable}, explanations that encompassed context and history of past successful interactions were able to improve failure identification and failure. 
Their context-based including history corresponds to our high-level explanation. 
On the other hand, 
with regard to our first research question, we have the following hypotheses:

\begin{itemize}[leftmargin=*]
    \item \textit{H1a: participants show better performance e.g. shorter task resolution time and successfully resolving the failure in the high explanation level compared to low and mid-levels.}
    
    \item \textit{H1b: participants are more satisfied when given more detailed explanations compared to lower or intermediate explanations.}
    
\end{itemize} 
Given our second research question, we hypothesize: 
\begin{itemize}[leftmargin=*]

    \item \textit{H2a: In final rounds, participants' performance and satisfaction in decaying conditions (with low explanations) is better than the fixed-low explanation condition. }   
    \item \textit{H2b: In final rounds, participants have comparable performance and satisfaction in decaying conditions (with low explanations) compared to fixed-high explanation conditions (with high explanations).}
\end{itemize}

For H2a, we specifically focus on Low-level explanations in round 3 and
expect participants to perform better in the decaying conditions (D1, D2) compared to the fixed low explanation condition C1 as they were given higher explanations in the previous rounds.
We also compare the performance in round 3 of C3 with (D1, D2) for which we expect participants to have similar perceptions and performances compared to C3, despite the low level of explanation, as they have already been exposed to higher explanations in earlier rounds (H2b).
The level of explanation is a between-subject variable for fixed strategy conditions (C1, C2, and C3).  
The level of explanation also varies within the decaying strategy conditions (D1 and D2) as it changes both between conditions and within the decaying conditions. 
The dependent variables are participants' performance in the task and their explanation satisfaction rating.
\begin{figure*}[hb]
    \subfloat[Carry failure success rate]{\includegraphics[width=.23\linewidth,trim={3.1cm 7.0cm 4.1cm 8.0cm},clip]{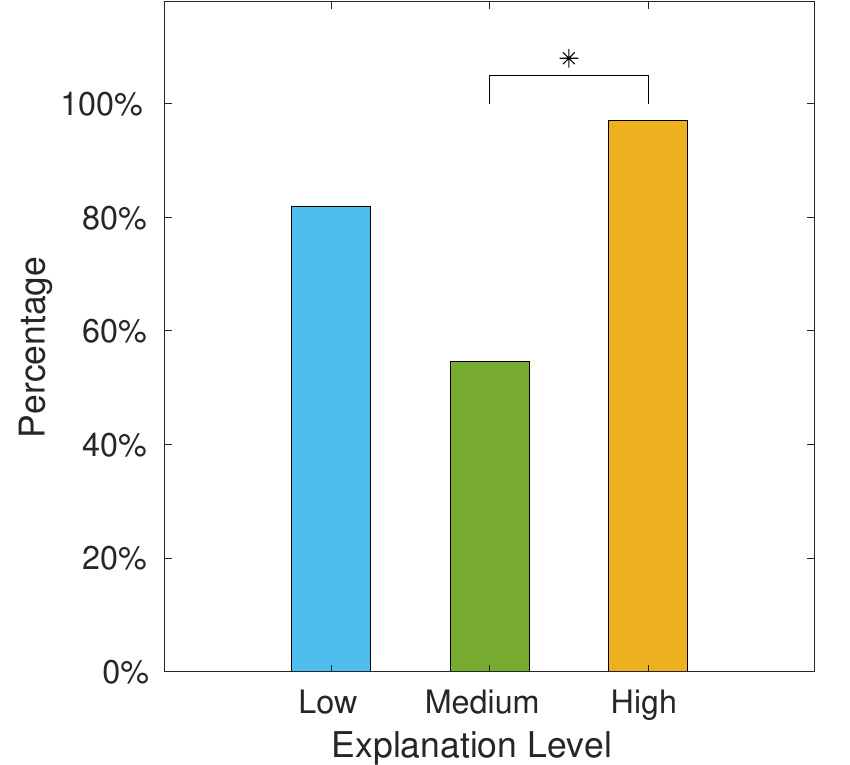}  
     }
     \subfloat[Carry failure resolution time]
     {\includegraphics[width=.23\linewidth,trim={3.1cm 7.0cm 4.1cm 8.0cm},clip]{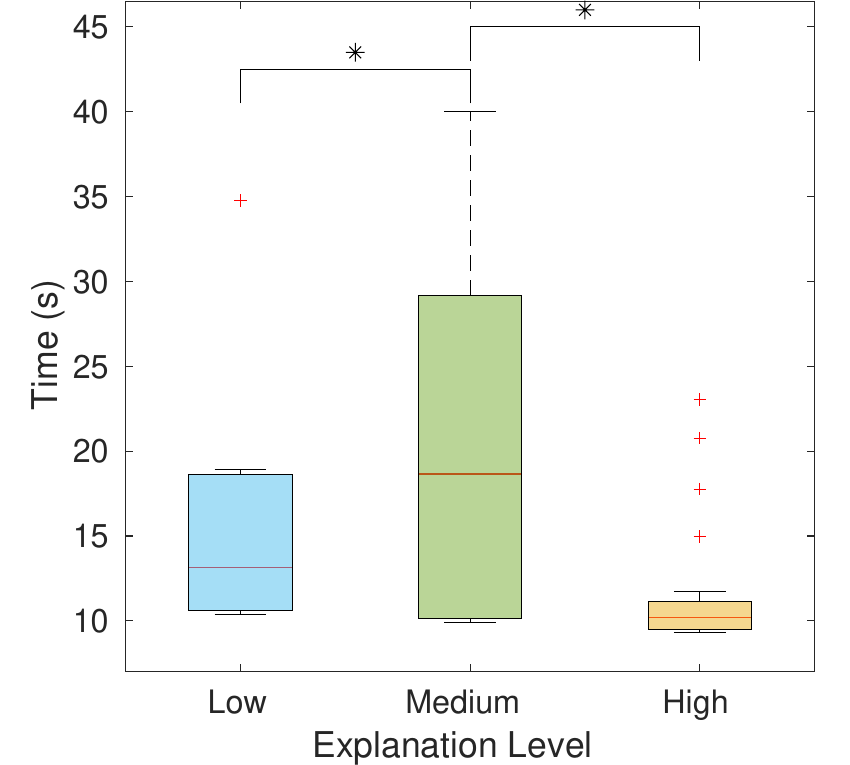}  
     }
      \subfloat[Place failure success rate]{\includegraphics[width=.23\linewidth,trim={3.1cm 7.0cm 4.1cm 8.0cm},clip]{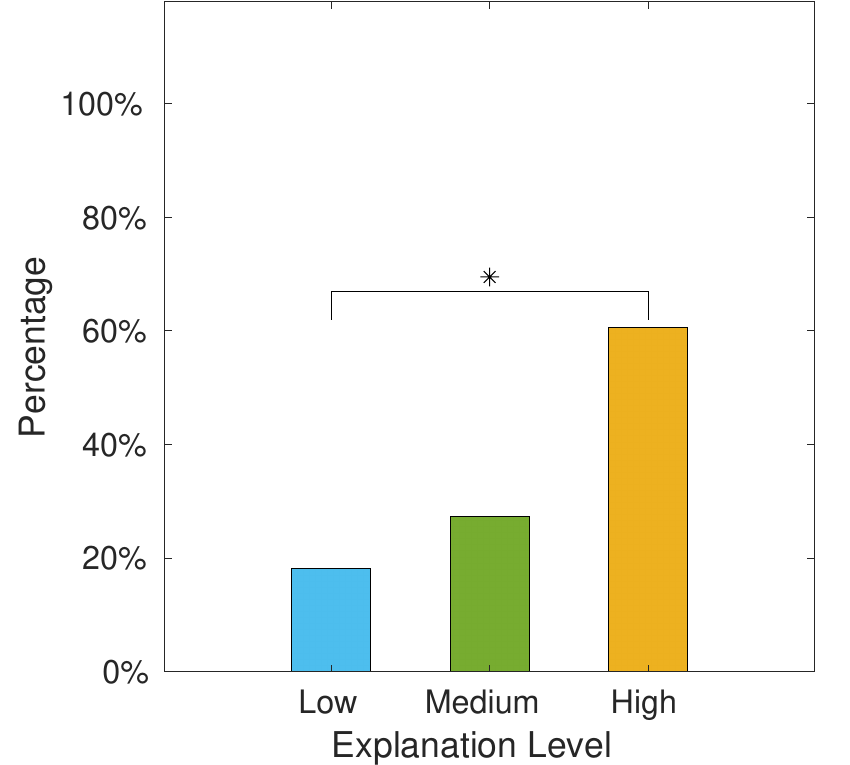}
     }
     \subfloat[Place failure resolution time]{\includegraphics[width=.23\linewidth,trim={3.1cm 7.0cm 4.1cm 7.85cm},clip]{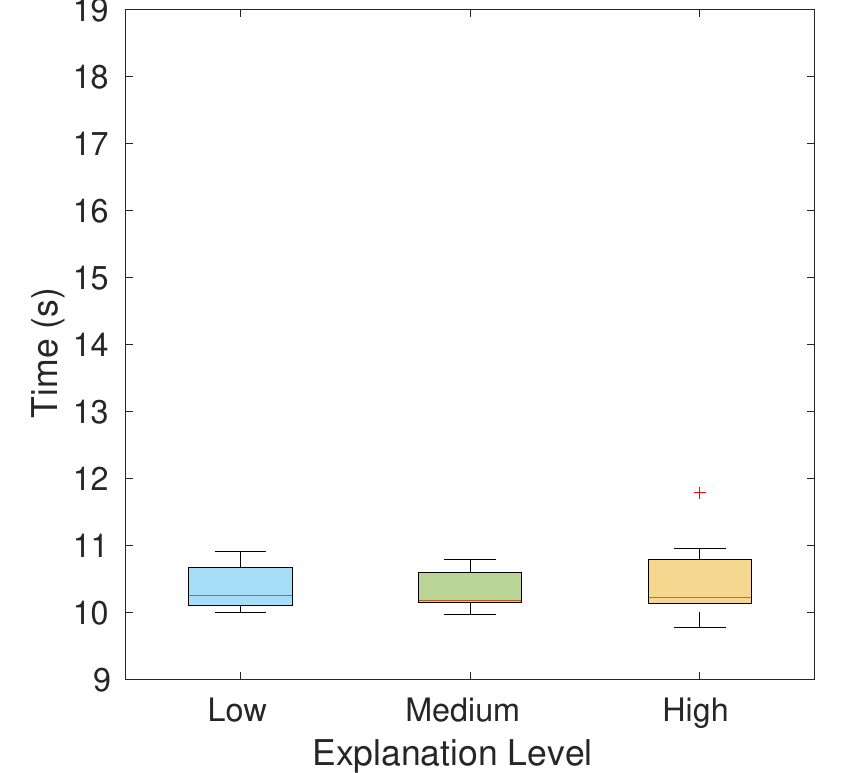}  
     }
\caption{Performance in terms of success rate and resolution time for round 1}
\label{fig:Round1-time and success}
\end{figure*}
\subsection{Measures}

In line with our hypotheses, the measures for this experiment were \textit{participants' performance} and \textit{participants' satisfaction}, collected through multiple variables and after the completion of each round of interaction. 

\subsubsection{Participants' performance}
We measure the performance over two dimensions corresponding to the instances where failure occurs, 
1) the time they take to intervene and resolve a failure, 
2) their success rate in resolving the failure, e.g. placing the object on the shelf. 

\textbf{Failure resolution time:}
$T_{res}$ is calculated from when the robot completes the explanation statement to when the participant completes the resolution, e.g. placing the object on the shelf. 

\textbf{Success rate of failure resolution:}
This is a measure of the successful resolution of each failure and it is measured differently depending on the type of failure as presented in the Collaborative Task Design section.

\subsubsection{Participants perception} The participant's perception was measured using an explanation satisfaction survey and some task-related questions. 
The task-related questions included more open-ended questions, designed to understand participants' approaches to resolving the failure beyond the robot's explanations. 
We are not reporting the qualitative analyses of the responses in this paper.  

\begin{table*}[t]
\vspace{2mm}
\setlength\abovecaptionskip{-0.3\baselineskip}
 \caption{Success rate in failure resolution}%
  \label{tab:failure_resolution_rate}%
  \centering
  \begingroup
\setlength{\tabcolsep}{2pt}
  \subfloat[][Round 1]{
\centering
\resizebox{0.55\columnwidth}{!}{
\begin{tabular}{c c c c}
\toprule
  Explanation & \multicolumn{3}{c}{Failures} \\ \cmidrule(lr){2-4} 
  Level & Pick & Carry & Place\\\midrule
\textbf{Low} & 100\% & 81.82\% & 18.18\% \\
\textbf{Medium} & 100\% & 54.55\% & 27.27\% \\
\textbf{High} & 100\% & 96.97\% & 60.61\%  \\

\hline
\end{tabular}
}
  }%
  \subfloat[][Round 3]{
\centering
\label{tab:succes_round3}
\resizebox{0.7\columnwidth}{!}{

\begin{tabular}{c c c c c}
\toprule
  Explanation & Explanation &\multicolumn{3}{c}{Failures} \\ \cmidrule(lr){3-5} 
  Strategy & Level & Pick & Carry & Place\\\midrule
\textbf{C1} & Low & 100\% & 100\% & 18.18\% \\
\textbf{C2} & Medium & 100\%  & 100\% & 45.45\%\\
\textbf{C3} & High & 100\%  & 100\%  & 72.73\%\\
\textbf{D1} & Low & 100\%  & 72.72\% & 18.18\%\\
\textbf{D2} & Low & 100\%  & 81.82\% & 36.36\%\\
\hline
\end{tabular}
}
  }%
  \subfloat[][Round 4]{
\centering
\resizebox{0.7\columnwidth}{!}{
\begin{tabular}{c c c c c}
\toprule
  Explanation & Explanation &\multicolumn{3}{c}{Failures} \\ \cmidrule(lr){3-5} 
  Strategy & Level & Pick & Carry & Place\\\midrule
\textbf{C1} & Low & 100\% & 90.91\% & 18.18\% \\
\textbf{C2} & Medium & 100\%  & 100\% & 36.36\%\\
\textbf{C3} & High & 100\%  & 100\%  & 72.73\%\\
\textbf{D1} & Low & 100\%  & 100\% & 45.45\%\\
\textbf{D2} & Low & 100\%  & 90.91\% & 63.63\%\\
\hline
\end{tabular}
}
  }
 \vspace{-2mm}
  \endgroup
\end{table*}

\textbf{Explanation satisfaction scale:}
To measure how participants were satisfied with the explanations at each round, we asked them to respond to 8 questions after completing each round. 
The questions were originally introduced and evaluated in \cite{hoffman2018metrics}.
They define explanation satisfaction as ``the degree to which users feel that they understand  the AI system or process being explained to them''. 
The questions were derived from the psychological literature on explanation and include several key attributes of explanations: \textit{understandability}, \textit{feeling of satisfaction}, \textit{sufficiency of detail}, \textit{completeness}, \textit{usefulness}, \textit{accuracy}, \textit{trustworthiness}. 

\subsection{Participants and Procedure}
We recruited sixty-nine participants via advertisement on campus. 
Our main criterion was that the participants had no prior experience in physical collaboration with a robot. 
Twelve participants had to be excluded from the analysis due to unaccounted robot failures beyond the failures designed for the experiment. 
The final sample size was N = 55 ($M = 26.63, SD =  7.42$) (21 Female, 33 Male, 1 Other) resulting in 11 participants per condition. 
At the start, the participants filled out the consent form for data and video collection and reading procedural instructions.
They were briefed about their role to place objects on the table and the robot's role to pick them up and place them on the shelf; however, no mention of the possible failures and related resolutions was presented.
After the completion of the experiment, they were given a debriefing sheet describing the aim of the study. 


\section{Results}

To prepare the data, first, we evaluated the internal consistency of the questionnaires using Cronbach’s alpha. 
The \textit{explanation satisfaction} questionnaire presented high internal consistency with Cronbach’s $\alpha = 0.79$, $\alpha = 0.91$, $\alpha = 0.92$, and $\alpha = 0.92$ for each round, respectively. 
\subsection{Impact of Explanation Level}
To investigate \textit{H1a} and \textit{H1b}, we only looked at the first round of interaction and grouped participants into groups of low, mid, and high explanation levels.  
This implied grouping the participants in conditions C3, D1, and D2 into \textit{High-level}, C2 into \textit{Mid-level}, and C1 into \textit{Low-level}. 
This decision was made to get a baseline for the explanation levels, additionally, to analyze strategies we need multiple rounds of interaction which we address in the next section. 

Given that each failure type required a different resolution and intervention to resolve that failure successfully, we analyzed the performances separately for each failure type.
Table \ref{tab:failure_resolution_rate}(a) shows the success rate in resolving the failures for each failure type in all three levels. 
For carry failures, Fisher's exact test $p = 0.0023$ shows a significant difference between the low, mid, and high explanation levels in successfully resolving the failure (Fig. \ref{fig:Round1-time and success}a).
According to post hoc tests $p = 0.0022$ this difference is significant between \textit{High-level} compared to the \textit{Mid-level}.
For place failure (Fig \ref{fig:Round1-time and success}c), according to Fisher's exact test $p = 0.0339$, participants that received the \textit{High-level} explanation were significantly more successful than the ones receiving the \textit{Low-level} explanation.


For our second measure of analyzing performance, we looked at the time participants took to resolve failure cases.
For pick and place failures, we observed no significant difference in the resolution times based on the explanation levels. 
For carry failures, Kruskal-Wallis chi-squared test $H(2) = x, p = 0.0075$ indicated that the resolution time significantly differed based on the explanation level.
Post hoc tests and Figure \ref{fig:Round1-time and success}b show the difference is significant between Low-level and Mid-level $p=0.0061$, and Mid-level and High-level $p=0.045$. 

\textbf{Results for \textit{H1a}:} Overall, the results partially support \textit{H1a}, where we expected participants to perform better in High-level explanations compared to Mid and Low-level.
However, the analyses show that failure type and how much the immediate resolution could be inferred from the environment, irrespective of the explanation, are important factors in the participants' performance. 
\begin{figure*}[hb]

\includegraphics[width=.07\linewidth,trim={9.5cm -6.0cm 8cm 2.0cm},clip]{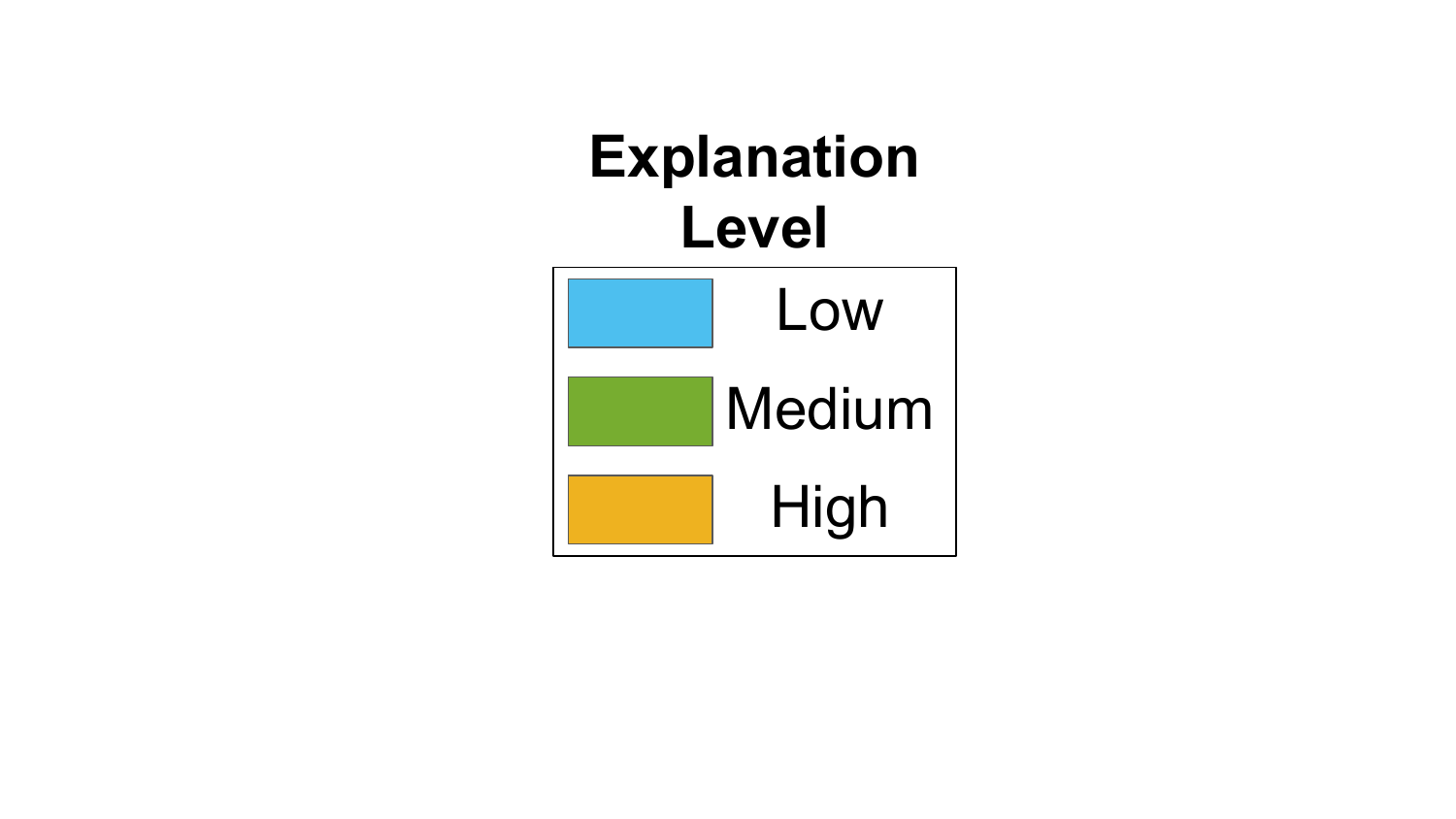}\hspace{-0.5em}
    \subfloat[Carry failure success rate]{\includegraphics[width=.23\linewidth,trim={3.1cm 7.0cm 4.1cm 8.0cm},clip]{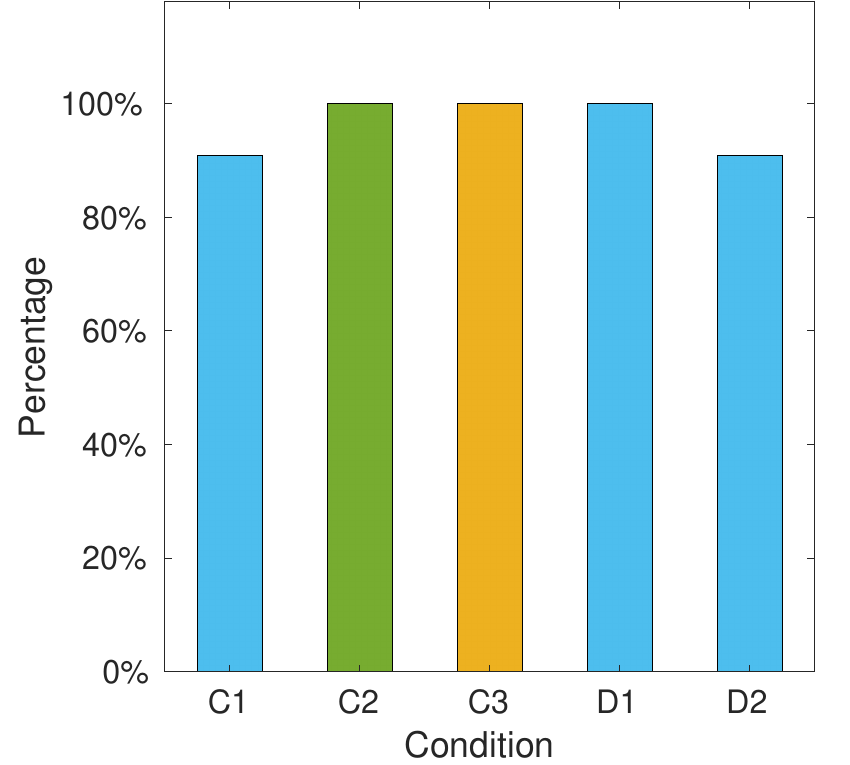}  \hspace{-0.5em}
     }
     \subfloat[Carry failure resolution time]{\includegraphics[width=.23\linewidth,trim={3.1cm 7.0cm 4.1cm 8.0cm},clip]{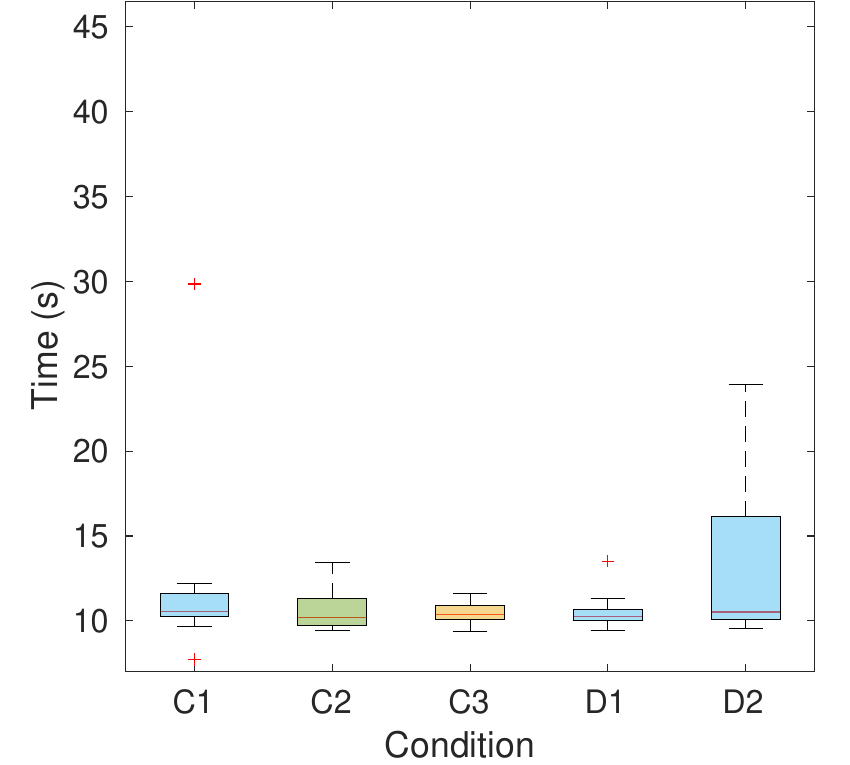}  \hspace{-0.5em}
     }
      \subfloat[Place failure success rate]{\includegraphics[width=.23\linewidth,trim={3.1cm 7.0cm 4.1cm 8.0cm},clip]{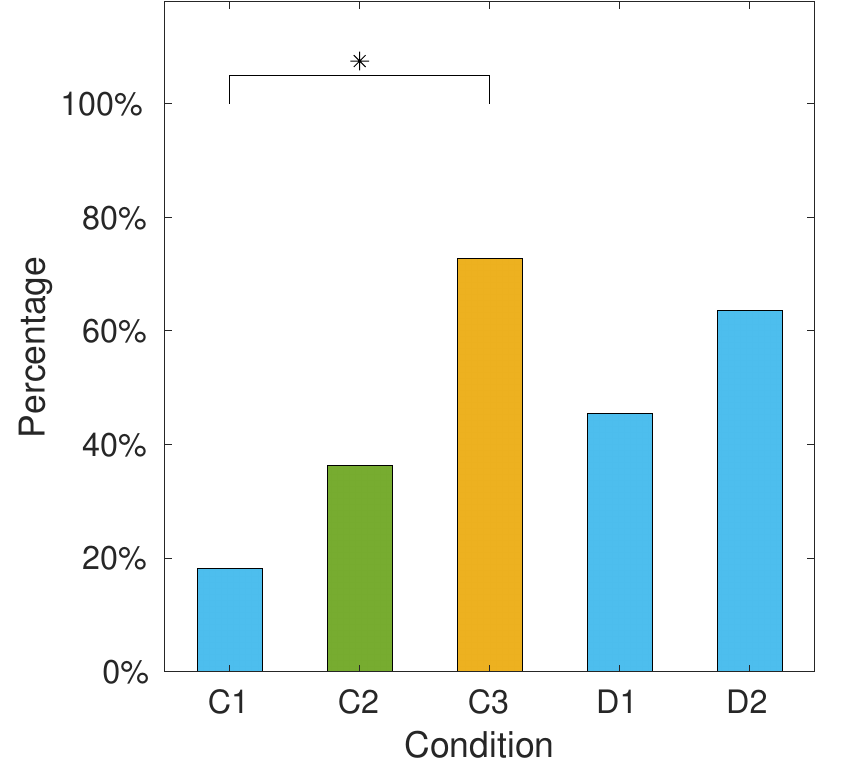}  \hspace{-0.5em}
     }
     \subfloat[Place failure resolution time]{\includegraphics[width=.23\linewidth,trim={3.1cm 7.0cm 4.1cm 7.85cm},clip]{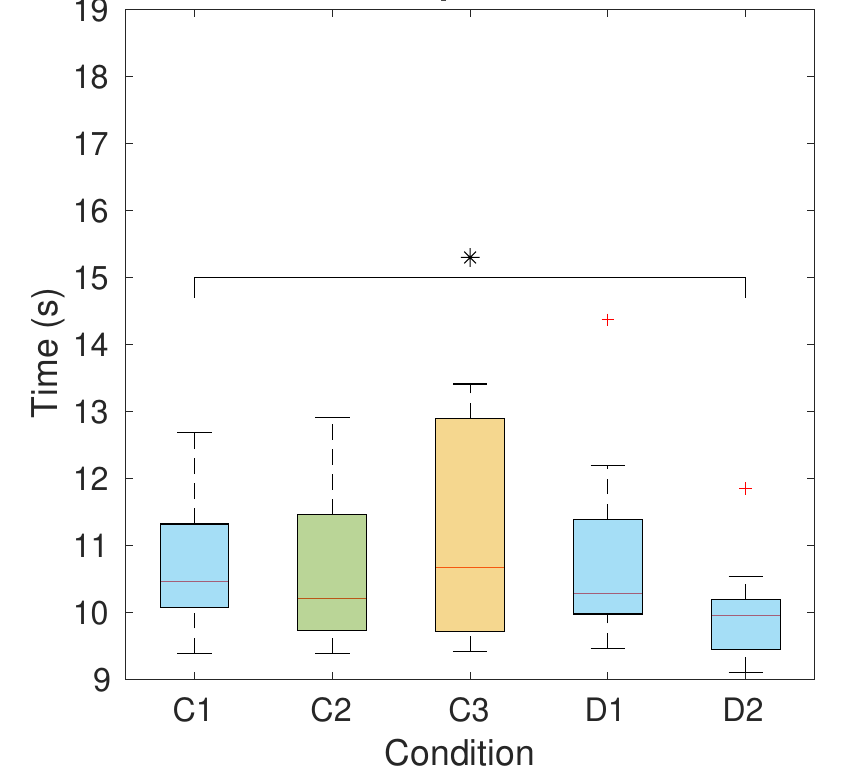}  \hspace{-0.5em}
     }
     \caption{Performance in terms of success rate and resolution time for round 3}
    \label{fig:round3}
    \vspace{-5mm}
\end{figure*}

\begin{figure*}[hb]

    \includegraphics[width=.07\linewidth,trim={9.5cm -6.0cm 8cm 2.0cm},clip]{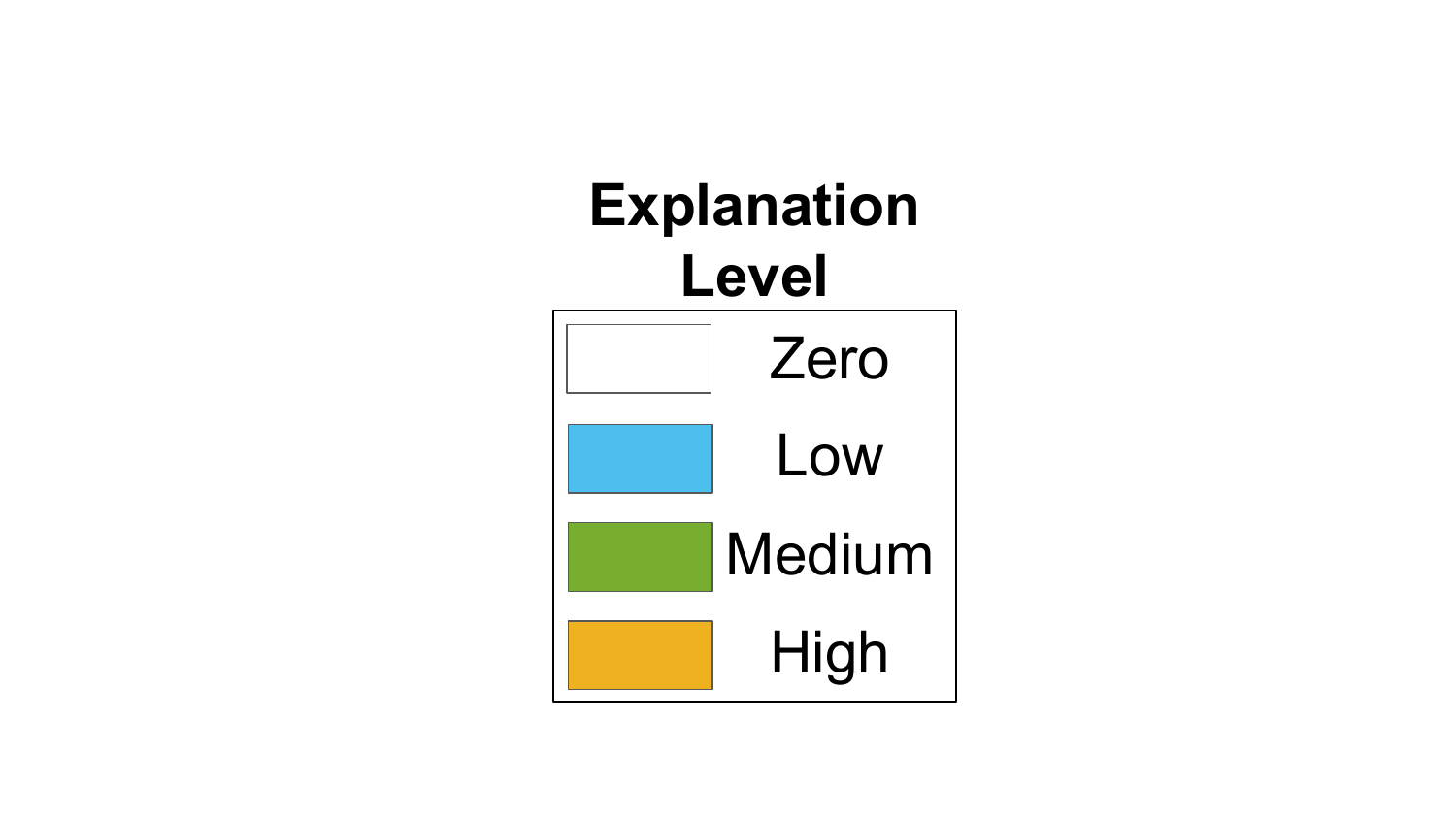} \hspace{-0.5em}\subfloat[Carry failure success rate]{\includegraphics[width=.23\linewidth,trim={3.1cm 7.0cm 4.1cm 8.0cm},clip]{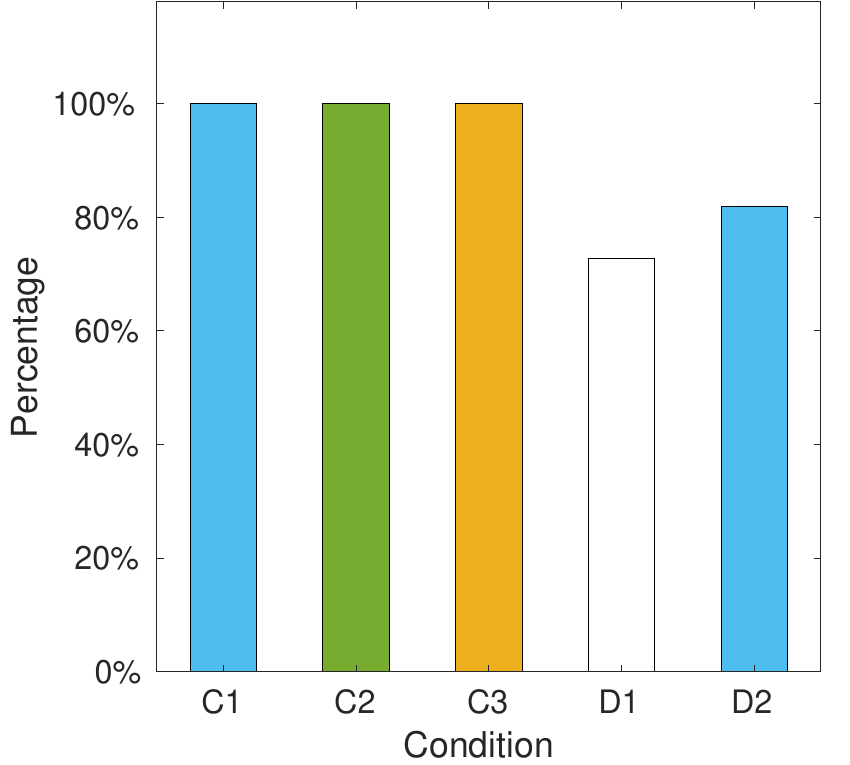}  \hspace{-0.5em}
     }
     \subfloat[Carry failure resolution time]{\includegraphics[width=.23\linewidth,trim={3.1cm 7.0cm 4.1cm 8.0cm},clip]{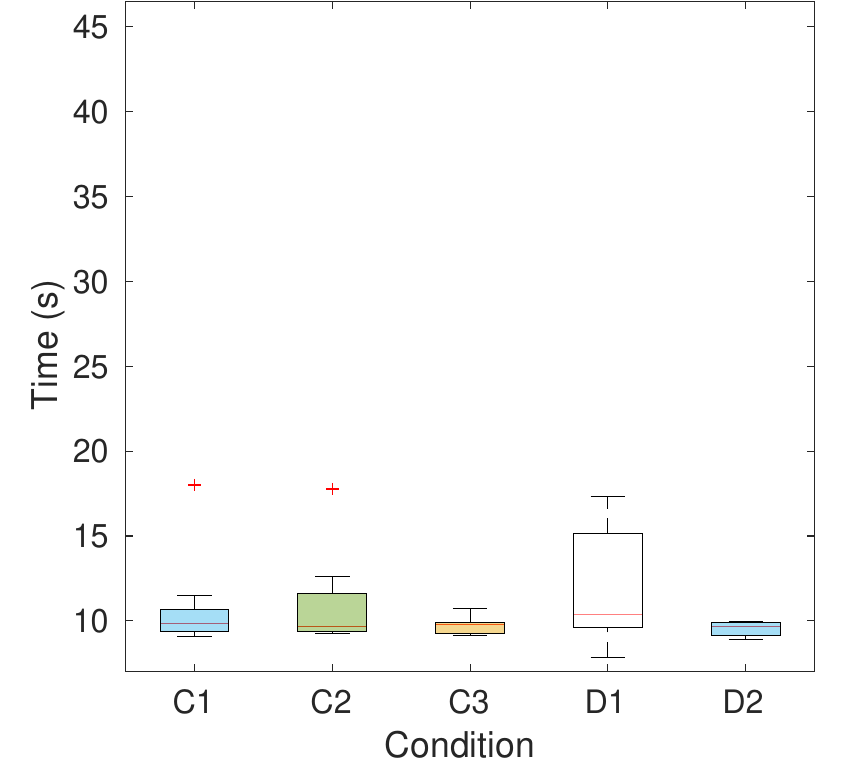}  \hspace{-0.5em}%
     }
      \subfloat[Place failure success rate]{\includegraphics[width=.23\linewidth,trim={3.1cm 7.0cm 4.1cm 8.0cm},clip]{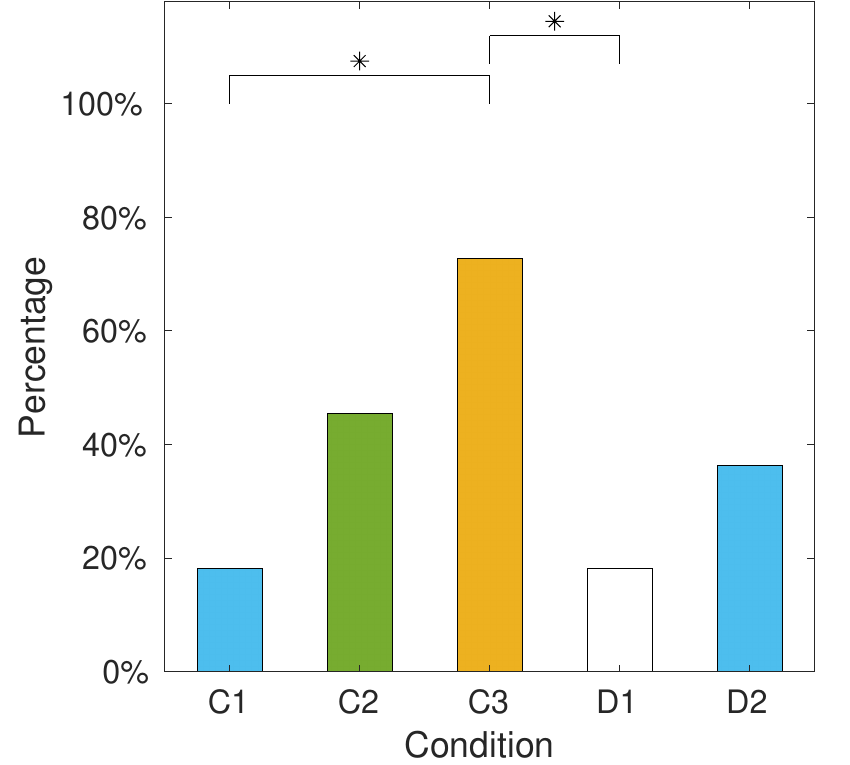}  \hspace{-0.5em}
     }
     \subfloat[Place failure resolution time]{\includegraphics[width=.23\linewidth,trim={3.1cm 7.0cm 4.1cm 7.85cm},clip]{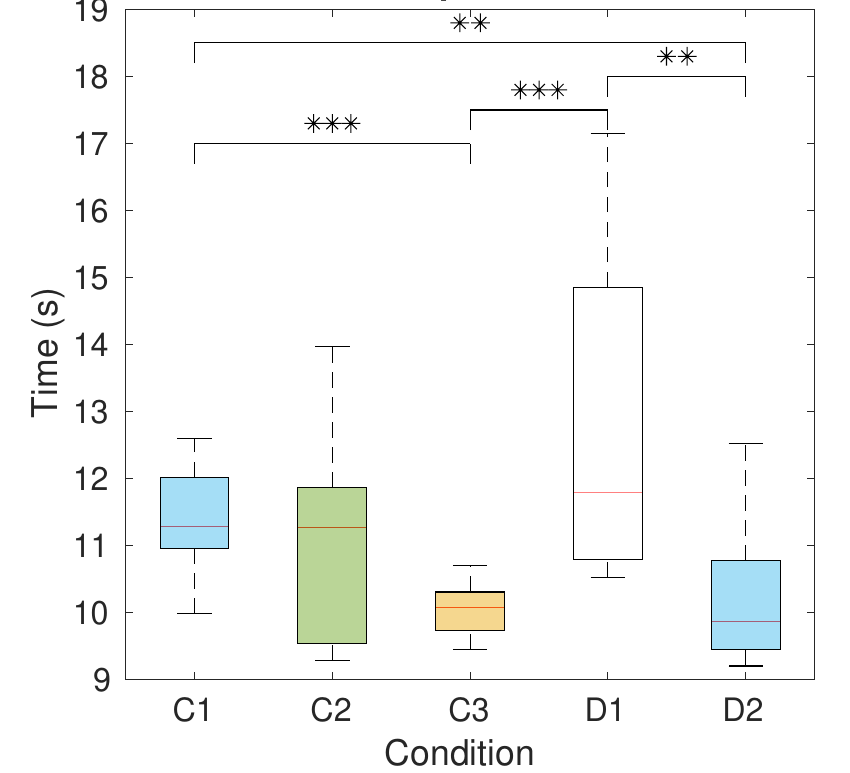}  
     }
     
    \caption{Performance in terms of success rate and resolution time for round 4}
     \label{fig:round4}
\end{figure*}

\begin{figure*}[tb]
\vspace{-2.5mm}
    \centering 
     \subfloat[Rating in round 1]{
     \includegraphics[width=.21\linewidth]{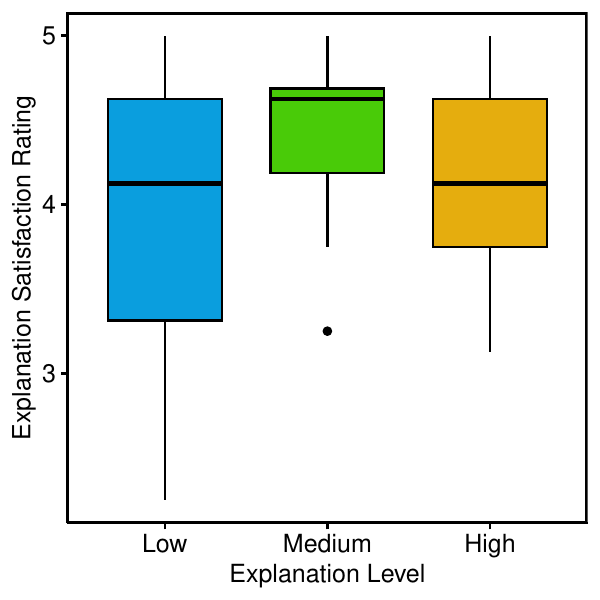}  
     }
     \subfloat[Rating in round 3]{
     \includegraphics[width=.21\linewidth]{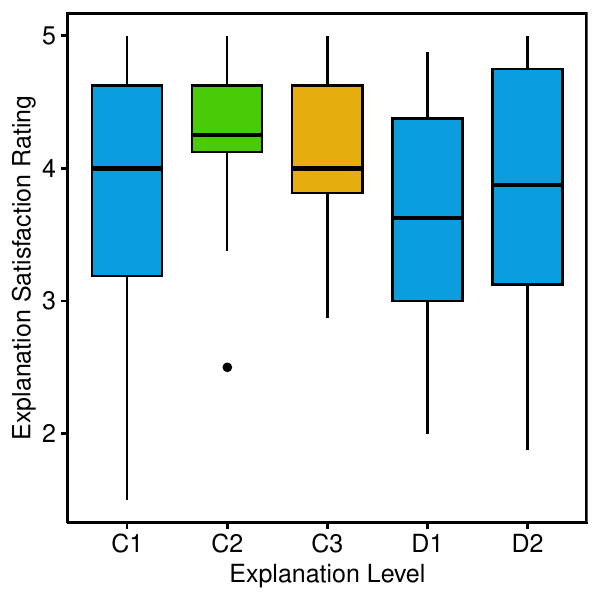}  
     }
     \subfloat[Ratings in round 4]{
     \includegraphics[width=.21\linewidth]{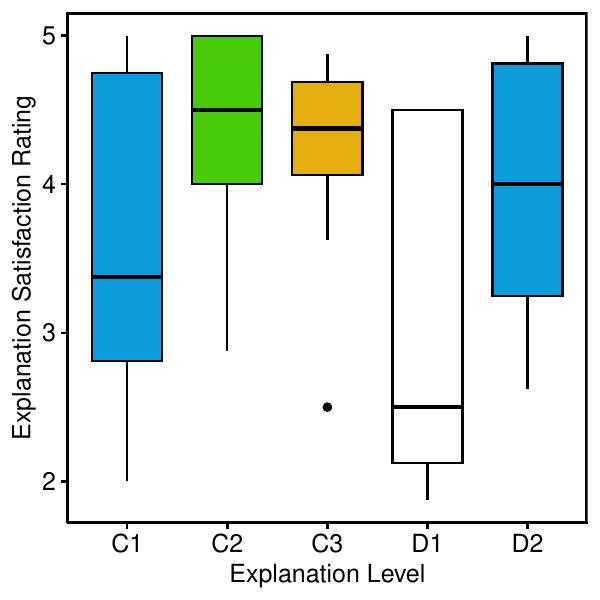}  
     }
     \subfloat[Rating trend for all rounds]
     {\includegraphics[width=.21\linewidth]{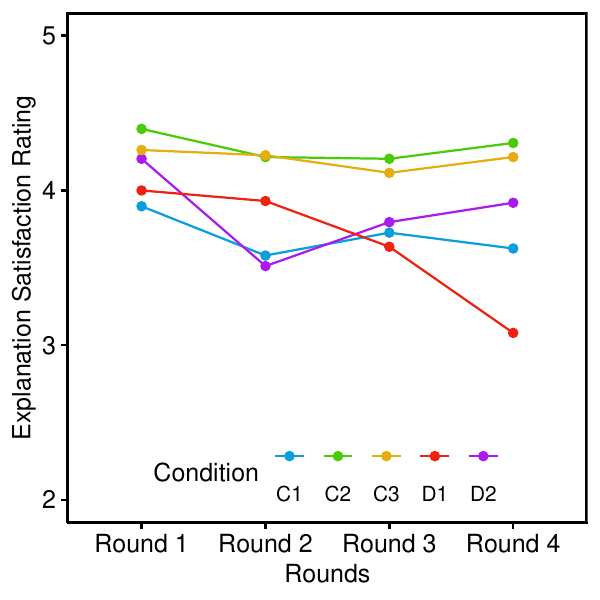}  
     }
     \caption{Explanation satisfaction ratings for all rounds}
      \label{fig:Round1-satisfaction}
\end{figure*}

\textbf{Results for \textit{H1b}:} Regarding \textit{H1b}, we analyzed participants' responses to the explanation satisfaction questionnaire after the first round.
 Kruskal-Wallis chi-squared test indicated no significant difference in the explanation satisfaction between the explanation levels $\textstyle{H(2) = 2.47, p = 0.2903 }$, rejecting our hypothesis.
The distribution of the satisfaction rating in round 1 is presented in Fig.~\ref{fig:Round1-satisfaction}a.

\subsection{Impact of Explanation Strategy}
To analyze the impact of the explanation strategy, we looked at participants' performance and satisfaction ratings in rounds 3 and 4 for conditions C1, C2, C3, D1, and D2. 
In \textit{H2a}, we are comparing the final round performances in the \textit{decaying} conditions, i.e. D1 and D2, versus the \textit{fixed} condition C1. 
In round three of these conditions, participants are receiving Low-level explanations with different a priori.
In round four, participants are receiving Low-level explanations in conditions C1 and D2, and Zero-level explanations in D1.

The percentages for the success rates in rounds 3 and round 4 are presented in Table \ref{tab:failure_resolution_rate}-(b),(c). 
For pick and carry failures, participants showed success rates above $80\%$ in all conditions. 
For place failures, while we observed better performances in D1 and D2 conditions compared to C1 as shown in Fig. \ref{fig:round3}c, the difference was not significantly different. 
Regarding the failure resolution times in round 3, no significant difference was observed for pick and carry failures between C1, D1, and D2 conditions.
However, for place failures, Kruskal-Wallis chi-squared test indicated that there was an overall difference in the resolution times between the three conditions $H(2) = 2.47, p = 0.2903 $.
The pairwise comparison confirmed that this difference was significant between C1 and D2 conditions $H(2) = 2.47, p = 0.0386$.

Furthermore, we explored the data in round 4, where the explanation level for condition D1 was reduced to baseline or none. 
As shown in Fig. \ref{fig:round4}, the performances for the D1 condition have decreased for all failure types,  with a significant difference for place failure cases.
By only looking at the performances for place failures in condition D1, we observe that after three rounds of interaction, participants were still not ready to resolve the failures without any explanation. 
Furthermore, the explanation satisfaction ratings for rounds 3 and 4 are presented in Fig. \ref{fig:Round1-satisfaction}b, and \ref{fig:Round1-satisfaction}c, and show no significant difference between the discussed conditions. 

\textbf{Results for \textit{H2a} and \textit{H2b}:} Overall, based on the performance and satisfaction results in the last rounds, we reject \textit{H2a}. 
However, we can accept \textit{H2b}, proving that participants in the last rounds of decaying explanation conditions; i.e. D1 and D2, showed comparable performances to the fixed-high explanation, i.e. C3. 

\section{Discussion}
\subsection{Impact of Explanation Level}
We observed that there is a significant effect of explanation level on the participants' performance.
Participants showed an overall higher success rate in resolving failures when given context-based high-level explanations with the history of past successful actions which was also evident from the shorter time in resolving the failures and completing the task. 
This is aligned with the results from \cite{das2021explainable}, where participants watched videos of the failures and respective explanations and their performance was evaluated based on success in identifying the cause of the failure and its resolution. 
Nevertheless, we noticed that the results are not generalizable over different types of failures.
For example, participants have shown above $80\%$ success rates in resolving \textit{pick failures} irrespective of the given level of explanation.
One reason lies in the nature of the failure to pick an object, which regardless of its cause, e.g. size, shape, or slippery edges, can be easily detected by a collaborator. 

On the other hand, the performances in resolving carry and place failures exhibited some significant differences based on explanation level. 
We identify that in \textit{carry failure} cases, the cause was not explicit, i.e. object weight was beyond the robot arm's limit. 
However, in contrast to our expectations, participants in \textit{Mid-level} conditions, had the worst performances compared to \textit{Low} and \textit{High-levels} (Fig. \ref{fig:Round1-time and success}a, \ref{fig:Round1-time and success}b).
This finding contributes to the argument that giving additional information without pointing to a cause or resolution can hinder human performance which is also aligned with Thagard's theory of explanatory coherence \cite{thagard1989extending}, where people prefer simpler explanations with fewer causes and more general explanations.
In \textit{place failure} cases, we observed significant performance improvement with the increase of explanation (Fig. \ref{fig:Round1-time and success}c, \ref{fig:Round1-time and success}d).
Several factors could contribute to this, including the harder detection of the resolution without receiving the appropriate explanation. 
It is plausible that participants understood the robot's failure to place the object on the shelf, however, they missed the exact reason, i.e., the inaccessibility of the lower shelf, and managed to place the object on the upper shelf, which was not the goal.

Overall our findings guide us to further investigate factors such as \textit{failure type}, with respect to its severity and \textit{information availability} as critical factors in estimating the need for explanation and generating the appropriate explanation upon failure.
While the literature on robot failures and trust evaluation considers failure severity to be an important factor that influences trust \cite{garza2018failure}, we further observe that situational awareness \cite{endsley1995toward} and the information availability play an important role too.
As a result, we conclude that: 1) if people can understand the failure and its resolution from the onset of failure, their performance is not influenced by the amount of provided explanations, and 2) more explanation does not automatically lead to better performance. 

\subsection{Impact of Explanation Strategy}
To understand how different explanation strategies performed, we analyzed participants' performances in later rounds, i.e., 3 and 4. 
In round 3, conditions C1, D1, and D2 had low-level explanations with different prior explanation levels. 
We observed that for \textit{carry failures}, performances were not significantly impacted by the explanation strategy. 
At this point, participants were already familiarized with the cause of this type of failure and resolution, and given their quite high success rate in the first round, they just kept improving.
On the other hand, we noticed that for \textit{place failure}, where the \textit{High-level} explanation was crucial to understanding the resolution, having a prior High-level explanation in conditions D1 and D2 improved the success rate. 
Consequently, the same improvement was observed in completing the task in a shorter time which was significant between conditions C1 and D2.
Overall, we conclude that in a repeated interaction scenario, a user responds better to a low level of explanation after being exposed to a higher level of explanation in prior rounds. 
This presents a strong justification for explanation strategies that reduce the level of explanations which reduces the overall task completion time.
Considering the results in condition D1, which included a Zero-level explanation in round 4, we conclude that not only the rate of reducing the explanation is important, but also the baseline where the explanation level reduces to.

\subsection{Limitations and Future Work}
Due to the exploratory nature of the study, we limited the number of possible conditions via pilot testing. 
Nevertheless, testing 5 conditions with 55 participants restricted us from drawing firm conclusions. 
While we observed some trends in the satisfaction ratings, having more participants wille enable us to surpass participants' personal differences.
This study was the first step in identifying the variables involved in how non-expert users perceive explanations after robot failures and the findings help us to improve our understanding of robotic failures and explanations strategies. 
Next, we plan to isolate some of these variables to determine the optimal adaptation that leads to higher human satisfaction and performance. 
to better evaluate how humans perceive the explanation and what type of adaptation is needed. 
We are extending the research by analyzing the dataset composed of participants' behaviors from their participation in the study when encountering failures. 
We are aiming at using social cues to recognize if participants have detected the failures \cite{bremers2023using} and utilize that information in a closed-loop system to adapt the explanation in response to the human's reaction to the failure.
Furthermore, we plan to conduct more user studies, investigating the conditions showing high varieties in performance and satisfaction ratings in more detail, including a detailed comparison of conditions C1, D1, and D2. 

\section{Conclusion}
In this work, we investigate what levels of explanation and what explanation strategies in repeated interactions help non-experts to assist a robot to recover from failures in a collaborative task. 
We introduce two types of explanation strategies in the context of repeated interactions i.e. fixed and decaying 
and designed a collaborative task i.e., picking and placing objects from a table to shelves, where we incorporated three types of commonly occurring failures in such tasks. 
A user study with 55 participants evaluated three variations of the fixed and two variations of the decaying strategies, with failures in four rounds of interaction. 
The results portrayed a bigger picture of how participants' performances in resolving the failures and their satisfaction with the robot's explanation is a function of types of failure, level of explanation, and strategy. 
We observed, for failures with a more explicit resolution, the level of explanation did not influence participants' performance or satisfaction. 
However, for failures where the cause of the failure contributed to resolving it, performance in the task and satisfaction were directly impacted by the context of the explanation. 
With regard to explanation strategies, we noticed that specifically for complex failures that can be resolved with the high explanation, we can aim for decaying strategies, where we can avoid repetitions and reduce overall collaboration times.
However, more modalities could be incorporated to decide the reduction rates, e.g. success rate in the previous rounds. 






\section*{ACKNOWLEDGMENT}

This work was partially funded by Digital Futures Research Center and Vinnova Competence Center for Trustworthy Edge Computing Systems and Applications at KTH.



\bibliographystyle{IEEEtran}
\bibliography{biblio}

\end{document}